\documentclass[runningheads]{llncs}
\usepackage[T1]{fontenc}
\usepackage{graphicx}
\usepackage{amssymb}
\usepackage{amsmath}
\usepackage{wrapfig}
\usepackage{threeparttable}
\usepackage{color}
\usepackage{diagbox}
\usepackage{bm}
\usepackage{pifont}
\usepackage{textcomp}
\usepackage{stfloats}
\usepackage{url}
\usepackage{verbatim}
\usepackage{graphicx}
\usepackage{threeparttable}
\usepackage{booktabs}
\newcommand{\dtb}[2]{
    \begin{tabular}[c]{@{}c@{}} 
        #1 \\
        #2
    \end{tabular}
}

\begin{document}

\title{Memory Matching is not Enough: Jointly Improving Memory Matching and Decoding for Video Object Segmentation}


\author{Jintu Zheng\inst{1} \and
Yun Liang\thanks{Corresponding author}\inst{2} \and
Yuqing Zhang\inst{3} \and
Wanchao Su\inst{4} 
}
\authorrunning{Jintu et al.}

\institute{University of Chinese Academy of Sciences, Beijing, China\\
\email{zhengjintu22@mails.ucas.ac.cn}\\
\and
South China Agricultural University, GuangDong, China\\
\email{yliang@scau.edu.cn}
\and
Beijing University Of Technology, Beijing, China\\
\email{yuqingz@emails.bjut.edu.cn}
\and
Department of Human Centered Computing, Faculty of Information Technology, Monash University, Melbourne, Australia\\
\email{wanchao.su@monash.edu}
}
\maketitle 
\begin{abstract}
Memory-based video object segmentation methods model multiple objects over long temporal-spatial spans by establishing memory bank, which achieve the remarkable performance.
However, they struggle to overcome the false matching and are prone to lose critical information, resulting in confusion among different objects.
In this paper, we propose an effective approach which jointly improving the matching and decoding stages to alleviate the false matching issue.
For the memory matching stage, we present a cost aware mechanism that suppresses the slight errors for short-term memory and a shunted cross-scale matching for long-term memory which establish a wide filed matching spaces for various object scales.
For the readout decoding stage, we implement a compensatory mechanism aims at recovering the essential information where missing at the matching stage.
Our approach achieves the outstanding performance in several popular benchmarks (i.e., DAVIS 2016\&2017 Val (92.4\%\&88.1\%), and DAVIS 2017 Test (83.9\%)), and achieves 84.8\%\&84.6\% on YouTubeVOS 2018\&2019 Val.
\keywords{Video Object Segmentation \and False Matching Alleviation \and Compensatory Decoding}
\end{abstract}

\begin{figure*}[ht]
    \centering
    \includegraphics[width=\linewidth]{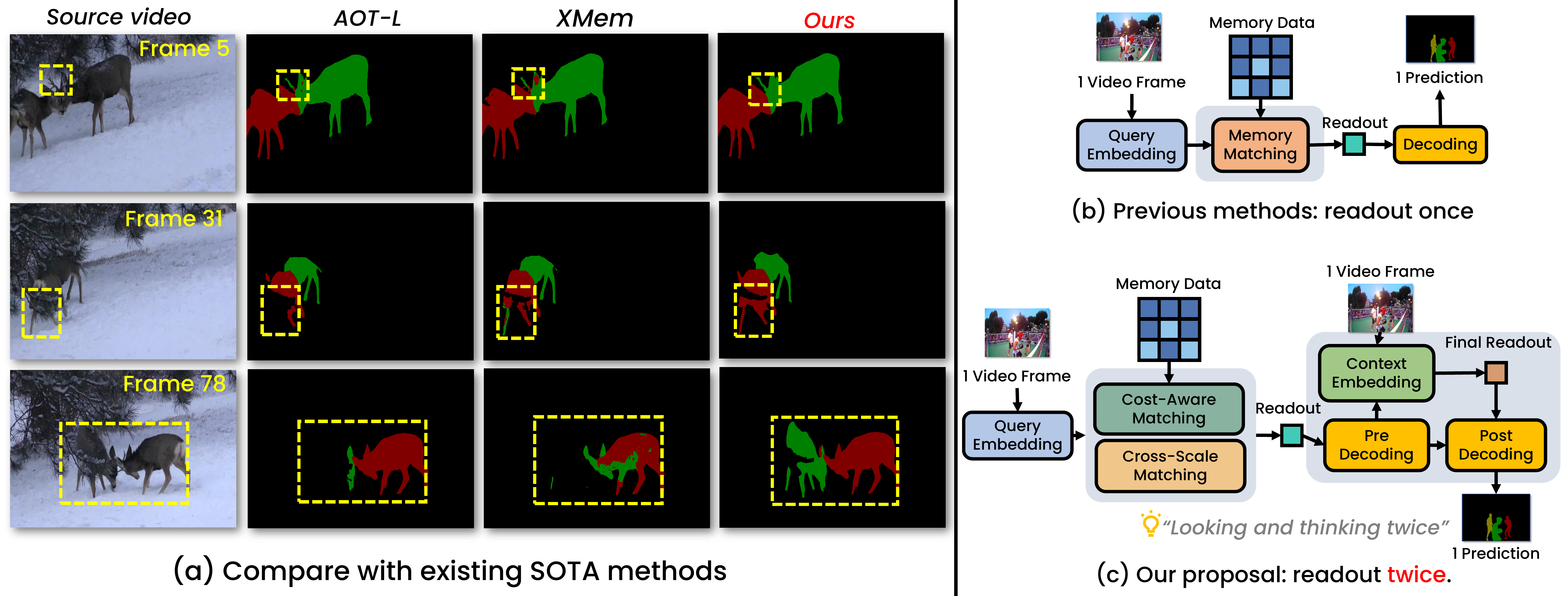}
    \caption{(a) Comparisons of a representative video clip in DAVIS 2017 Test. AOT \cite{AOT} and XMem \cite{XMem} (two state-of-the-art matching-based VOS models) present false matching errors and our method can produce more accurate masks. (b-c) A simplified comparison on pipeline between ours and previous matching-based methods. Previous matching-based lacks the consideration of combining two stages to improve.}
    \label{fig_banner}
\end{figure*}

\section{Introduction}
Video object segmentation (VOS) is a fundamental procedure for many multimedia applications, such as special effects editing in movies, robot interaction, and smart camera surveillance, which requires instance segmentation of objects of interest in videos.
The work in this paper focuses on semi-supervised VOS, which completes instance segmentation of the remaining frames based on multiple instances given in the first frame.
Recently, matching-based approaches have gained popularity, wherein the basic idea is to establishing and maintaining a memory to store previous frames and their corresponding masks.
These stored memories are then matched with the query frame to generate a memory readout, and finally produce masks of the target objects from the memory readout.

Memory matching essentially relates to the accuracy in generating the target object masks, which becomes a crucial component in improving the accuracy of VOS tasks.
Early matching-based methods such as STM \cite{STM} and its variants \cite{STCN,KMN,HMMN} employ attention mechanisms to achieve matching between query frames and the memory.
Such methods treat all the memory units with equal importance, without special design regarding the individual memory unit in the process.
Inspired by the human cognitive process where there is a distinctive difference between short-term and long-term memories, current memory matching methods treat short-term and long-term memory differently in VOS.
Long-term memory stores multiple historical frames, and records the change across frames in a coarse-grained manner.
The objects in long-term memory may have different scales, so the long-term matching mechanism should not be limited to a single-scale.
Short-term memory focuses on the adjacent frames, which are similar with each other, meaning the variations are fine-grained.
Thus, the short-term matching mechanism must capture the variants sufficiently.

Some state-of-the-art methods (i.e., AOT \cite{AOT}, variants of AOT \cite{AOST,DeAOT}, and XMem \cite{XMem}) divide long-term and short-term memory, these methods still have limitations:
As shown in Fig. \ref{fig_banner}(a), the AOT \cite{AOT} and XMem \cite{XMem} produce results with slight errors in the impact of the short-term memory insufficiency (Frame 5 in Fig. \ref{fig_banner}(a)), the later frames (Frame 31 \& 78 in Fig. \ref{fig_banner}(a)) present object confusion and crucial information loss.
On the one hand, they employ single-scale attention in long-term memory, which makes them exhibit rapid performance degradation in handling multiple objects, especially undergoing different morphological changes simultaneously.
On the other hand, these methods also need to improve in matching short-term memory.
For example, AOT \cite{AOT} and its variants \cite{AOST,DeAOT} implement the local correlation attention, which may lose critical information when the morphological changes occur outside the local memory unit's perceptive field.
In addition, slight errors in short-term memory matching accumulate in long-term memory, which can be fatal for VOS methods that rely on long-term memory.

In this paper, we propose an improved memory matching mechanism, including cost-aware matching for short-term memory and cross-scale matching for long-term memory.
Cost-aware matching mechanism focuses on a stronger relationship between corresponding pixels of adjacent frames.
Inspired by optical flow prediction methods \cite{pwc,Raft,flowformer}, we construct the cost volume for the query frame and the previous frame.
Cost volume is a vector that stores the matching degree of corresponding pixels between two frames \cite{costvolume}.
After patch embedding the cost volume, we introduce a group of learnable query tokens for collecting coarse-grained spatial variations.
Furthermore, to explore fine-grained details, we construct a spatial readout head via SS-attention \cite{twins}.
Note that cost volume in our cost-aware matching is a global relationship of pixels in adjacent frames which is different from the neighborhood correlation of short-term transformer in AOT \cite{AOT} and its variants \cite{AOST,DeAOT}.
We implement multiple scales for long-term memory and more effective models objects of various scales simultaneously within a matching block.
Our improved memory mechanism reduces the accumulation of slight errors in short-term memory and adequately adapts the objects of various spatial morphology in the previous frames.

Matching-based VOS methods inevitably produce false matches \cite{STM,STCN,XMem}, which may lead to object confusion (see XMem in Fig. \ref{fig_banner}(a)) or missing objects (see AOT in Fig. \ref{fig_banner}(a)).
However, matching-based methods usually focus solely on improving memory matching, and they implement a naive FPN \cite{FPN} for decoding (e.g., STCN \cite{STCN}, HMMN \cite{HMMN} and RMNet \cite{RMNet}), lacking consideration of modifying the decoding process.
AOT \cite{AOT} and XMem \cite{XMem} make extensive modifications for memory matching, the problem of false matches still exists.
Unlike fully supervised segmentation that understands rich semantics, semi-supervised VOS requires more the low-level semantic feature prompt of target objects.
We argue that suppressing false matches requires improving memory matching and improving decoding process.
There is significant potential for improving the decoding process.
For instance, AOML \cite{AOML} achieves excellent performance by designing bi-decoders for online learning VOS.
CFBI \cite{CFBI} and CFBI+ \cite{CFBI2} emphasizes separating the foreground and background in decodng process to improve matching.
They have an explicit foreground-background embedding feature and low-level feature incorporation, which are beneficial for distinguishing the foreground from the background.
Such explicit foreground-background distinction still struggle to overcome the false matching problem.
Our paper aims to give a more suitable and comprehensive answer, which jointly improves both stages and rethinks all details toward reducing the false matching instead of the simple foreground-background distinction.

Therefore, we propose a compensatory decoding mechanism (as shown in Fig. \ref{fig_banner}(c)), which consists of three steps, 1) pre-decoding, 2) context embedding, and 3) post-decoding.
The initial memory readout inevitably lose some critical information and looking twice at the original image effectively compensates such losses; thus, we embed a context embedding process in the decoding stage to force the encoder to look at the query frame one more time, which supplements the critical information lost in the memory matching stage.
Pre-decoding provides a guiding prompt for context embedding, and the post-decoding generates the final segmentation masks.
The compensatory decoding mechanism not only sufficiently embeds the critical information of the target objects but also suppresses the false matches in the initial memory readout to some extent.

The mainly contributions of this paper are summarized as:

\begin{itemize}
    \item Different existing methods, we improve the matching and decoding stages in a jointly paradigm, which give a more suitable and comprehensive answer to alleviate false matching issue.

    \item We propose a improved mechanisms for the memory matching stage. Cost-aware matching in short-term memory prompts the network to perceive changes between two frames more adequately. Cross-scale matching in long-term memory prompts the network to explore the variations in different scaled objects.

    \item We propose a novel compensatory decoding mechanism that can suppress false matches and supplement the crucial information loss of target objects for the readout decoding stage.

    \item Our approach achieves state-of-the-art performance in several popular benchmarks (i.e., DAVIS 2016\&2017 Val (92.4\%\&88.1\%), and DAVIS 2017 Test (83.9\%)), and achieves 84.8\%\&84.6\% on YouTubeVOS 2018\&2019 Val with the specific training strategy.
\end{itemize}  
\section{Related Work}
\noindent\textbf{Semi-supervised Video Object Segmentation}
In semi-supervised VOS, according to the object masks given in the first frame of the video, the corresponding object is segmented in the remaining video frames.
The mainstream of semi-supervised VOS methods can be roughly divided into online fine-tuning and methods without fine-tuning.
In the methods without fine-tuning, matching based is the popular study branch.

\noindent\textbf{Matching based Methods}
Matching-based is an VOS method without fine-tuning that has achieved notable success, and the proposal in this paper is focus on the matching based methods.
STM \cite{STM} introduces a spatio-temporal network to establish the matching relationship between the current frame (query frame) and all historical frames (memory), which can be roughly divided into three stages: query embedding, memory matching, and readout decoding.
The following matching-based methods almost focus on improving the memory matching stage.
Some methods \cite{LCM,EGMN,KMMN} focus on designing novel memory structures, such as HMMN \cite{HMMN}, which designs a hierarchical memory structure.
In addition, some methods \cite{STCN,KMN,RMNet} proposed novel matching mechanisms, such as STCN \cite{STCN} proposed utilizing negative squared euclidean distance to calculate the affinity of matching, and KMN \cite{KMN} introduced gaussian kernel to reduce the non-locality of matching.
All the above methods treat all memories fairly, which makes the performance of the memory model have a bottleneck.
Recently, some state-of-the-art methods such as AOT \cite{AOT} and its variants (AOST \cite{AOST}, DeAOT \cite{DeAOT}), XMem \cite{XMem} distinguish memory between long and short term to improve the inference performance.
However, there are still significant improvements possibilities for these methods.
On the one hand, the design of this long-short term matching mechanism does not fully adapt to multi-scale objects and ignores part of the crucial inter-frame changes.
In this paper, we propose an improved long short-term memory matching mechanism that outperforms these methods in performance.
On the other hand, we argue that \textit{"memory matching is not enough"}, and previous methods have neglected the effect of improvements on the readout decoding for semi-supervised VOS.


\begin{figure*}[http]
    \centering
    \includegraphics[width=\textwidth]{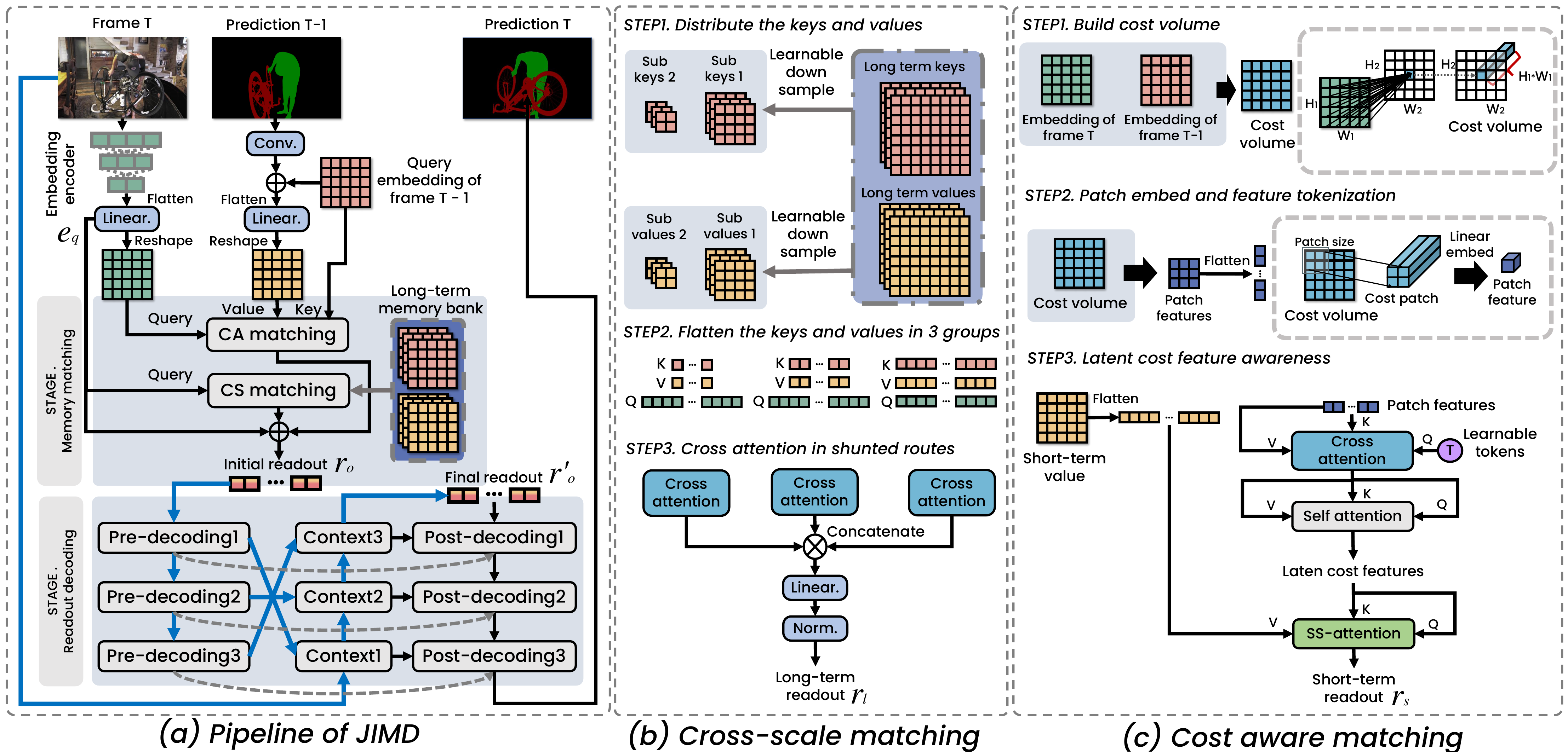} 
    \caption{(a) Pipeline of our proposal, which improves the memory matching stage by cost-aware and cross-scale matching, and improves the decoding stage by compensatory decoding. (b-c) Illustration of cross-scale and cost-aware matching.}
    \label{fig:ooview}
    \vspace{-5mm}
\end{figure*}

\section{Methodology}
\subsection{Proposal Overview}\label{sec:ooview}
We propose an effective approach to alleviate the false matching issue from the memory matching and readout decoding stages.
We name this proposal as \textbf{J}ointly \textbf{I}mprove \textbf{M}atching and \textbf{D}ecoding (JIMD) in the following content.
The pipeline of JIMD is illustrated in \ref{fig:ooview}.
JIMD sequentially processes each frame for a video clip.
For the current frame $T$, we extract the backbone features from the embedding encoder and then input into a linear layer to obtain the embedding query feature $e_q$.
Then we implement the cost volume matching for short-term memory, and employ the cross scale matching for the long-term memory to obtain two matching readout results (i.e., $r_s$ and $r_l$).
Initial readout feature $r_o$ can be formulated as: $r_o = r_l + r_s + e_q$.
We firstly decode the $r_o$ for guiding the context block to extract the information form source frame.
We obtain the final readout feature $r_o^{'}$ after embedding the source context into the initial readout $r_o$.
Finally, we decode the readout feature $r_o^{'}$ and upsample into the object masks.
During the entire procedure, JIMD maintains two sets of memory data, which are long-term and short-term memories.
The short-term memory is stacked into the long-term memory at intervals, and the memory data is stored as \emph{key} \& \emph{value}.
Memory value is generated by fusing the previous embedding feature and the mask feature from the ID module, which is borrowed from AOT \cite{AOT}.
Here, we implement a convolution with kernel size of $17\times 17$ and stride of 16 to encode the masks of frame $T-1$ as the identify encoder.

\subsection{Cross-Scale in Long-Term Matching}\label{sec:scale_matching}
Long-term memory records the change across frames in a coarse-grained manner (e.g., an object may have multiple morphologies across frames or multiple scale objects in the same frame).
Introducing cross-scale in long-term memory is beneficial to match targets with variable scales.
Therefore, we shunt the keys and values, downsampling the long-term memory keys and values at different scales.
Let the $K_{\{0,1,2,...,T-1\}}$ and $V_{\{0,1,2,...,T-1\}}$ denote all previous keys and values in long-term memory.
As shown in \ref{fig:ooview}, we employ three spatial rates for non-local matching.
The downsample is a convolution layer with the decreasing kernel size as illustrated in \ref{fig:ooview}.
Here, we denote $d_i$ as the spatial rate for downsampling:

\begin{footnotesize}
\begin{align}
K^{di}_{\{0,1,2,...,T-1\}} = downsample(K_{\{0,1,2,...,T-1\}}, d_i) \\
V^{di}_{\{0,1,2,...,T-1\}} = downsample(V_{\{0,1,2,...,T-1\}}, d_i)
\end{align}
\end{footnotesize}
We perform cross-attention with $e_q$ after obtaining keys and values of different sizes:

\begin{footnotesize}
\begin{equation}
\sigma_i = Atten(e_q, K^{d_i}_{\{0,1,2,...,T-1\}}, V^{d_i}_{\{0,1,2,...,T-1\}}).
\end{equation}
\end{footnotesize}
Three attention results are concatenated, then projected as the readout feature $r_l$ of long-term memory:

\begin{footnotesize}
\begin{equation}
r_l = LN(concat(\sigma_0, \sigma_1, \sigma_2)),
\end{equation}
\end{footnotesize}
where $LN$ is the linear norm layer.
This shunting matching benefits the long-term memory and performs well in handling multi-scale cases.

\subsection{Cost-Aware Matching}\label{sec:cost_matching}
The purpose of the short-term memory matching mechanism is to learn the changes in adjacent frames, which is crucial for VOS.
If the short-term memory matching produces inaccurate masks that are used in generating values for the next round, it would lead to a vicious circle process.
Representing the changes in adjacent frames is essential for effective short-term memory matching, it requires more larger local receptive field but with a low computational cost increasement.
We construct a cost volume to represent the interframe variations and tokenize the cost volume with patch embedding.
Then we implement the multi-head attention mechanism to produce the latent cost features that involve critical variations.
Finally, the SS-attention \cite{twins} generates the readout feature with the latent cost features and the previous frame's value.
We construct cost-aware matching as a transformer architecture, as shown in \ref{fig:ooview}.

\noindent\textbf{Cost Volume Generation.}
The first step of short-term memory matching requires the generation of initial information representing the variations between adjacent frames.
Instead of the local correlation matching in AOT \cite{AOT} and its variants \cite{AOST,DeAOT} with high computational cost, we build a global 4D cost volume $c_m\in\mathbb{R}^{H_1\times W_1\times H_2 \times W_2}$ for $e_q$ and $e_k$, the volume is constructed through dot product operations.
Here, $H_1$ and $W_1$ donate the $e_k$'s height and width, $H_2$ and $W_2$ donate the $e_q$'s height and width. 
As shown in the bottom right corner of \ref{fig:ooview}, each 2D sub-map in cost volume $c_m$ can be regarded as the visual similarity between the source pixel and all target pixels, and $e_k$ is derived from the $e_q$ of the previous frame.
We patch embed the cost volume, which divides the cost volume into multiple patches according to the patch size, and perform feature embedding on each patch, as shown in the upper right corner of \ref{fig:ooview}.
We define the number of the patches as $S$, embedding dimension as
$C^{\prime}$, the patch embedding features $P_s\in\mathbb{R}^{(H_1*W_1)\times S \times C^{\prime}}$ as:

\begin{footnotesize}
\begin{align}
P_s = PatchEmbed(e_q \centerdot e_k).
\end{align}
\end{footnotesize}

\noindent\textbf{Patch Features Tokenization.}
For complicated variations between adjacent frames, we need to pay more attention to the patches relating
to the target objects.
To achieve this goal, we introduce a set of learnable tokens $L_q\in\mathbb{R}^{N_l \times C_l}$ to extract the patch features $P_s^{\prime}\in\mathbb{R}^{(H_1*W_1)\times N_l \times C_l}$ that contains latent critical variations, where $N_l$ is the token number and $C_l$ is the tokens' embedding dimension.
We implement a cross-attention, which applies the learnable tokens $L_q$ as query, patch embedding features $P_s$ as the key and value:

\begin{footnotesize}
\begin{equation}
P_s^{\prime} = Atten(L_{q}, P_{s}, P_{s}).
\end{equation}
\end{footnotesize}
Then we apply a multi-head self-attention to output the latent cost features $C_{l}\in\mathbb{R}^{(H_1*W_1)\times N_l \times C_l}$.
The latent cost features $C_l$ implicitly represent critical variants between adjacent frames.

\noindent\textbf{Producing Readout Feature.}
We implement the SS-attention \cite{twins}, which focus more on capturing spatial features to
produce short-term readout result $r_s$: 

\begin{footnotesize}
\begin{equation}
r_s = SSAtten(C_{l}, C_{l}, V_{T-1}) + C_l,
\end{equation}
\end{footnotesize}

where $V_{T-1}$ is the frame $T-1$ value, and there is a final fusion for attention output and $C_l$.
After the cost-aware matching, the short-term readout feature $r_s$ sufficiently capture the fine-grained variations between adjacent frames.

\subsection{Compensatory Decoding}\label{sec:decoding}
\begin{wrapfigure}{r}{0.5\textwidth}
    \centering
    \vspace{-2em}
    \includegraphics[width=\linewidth]{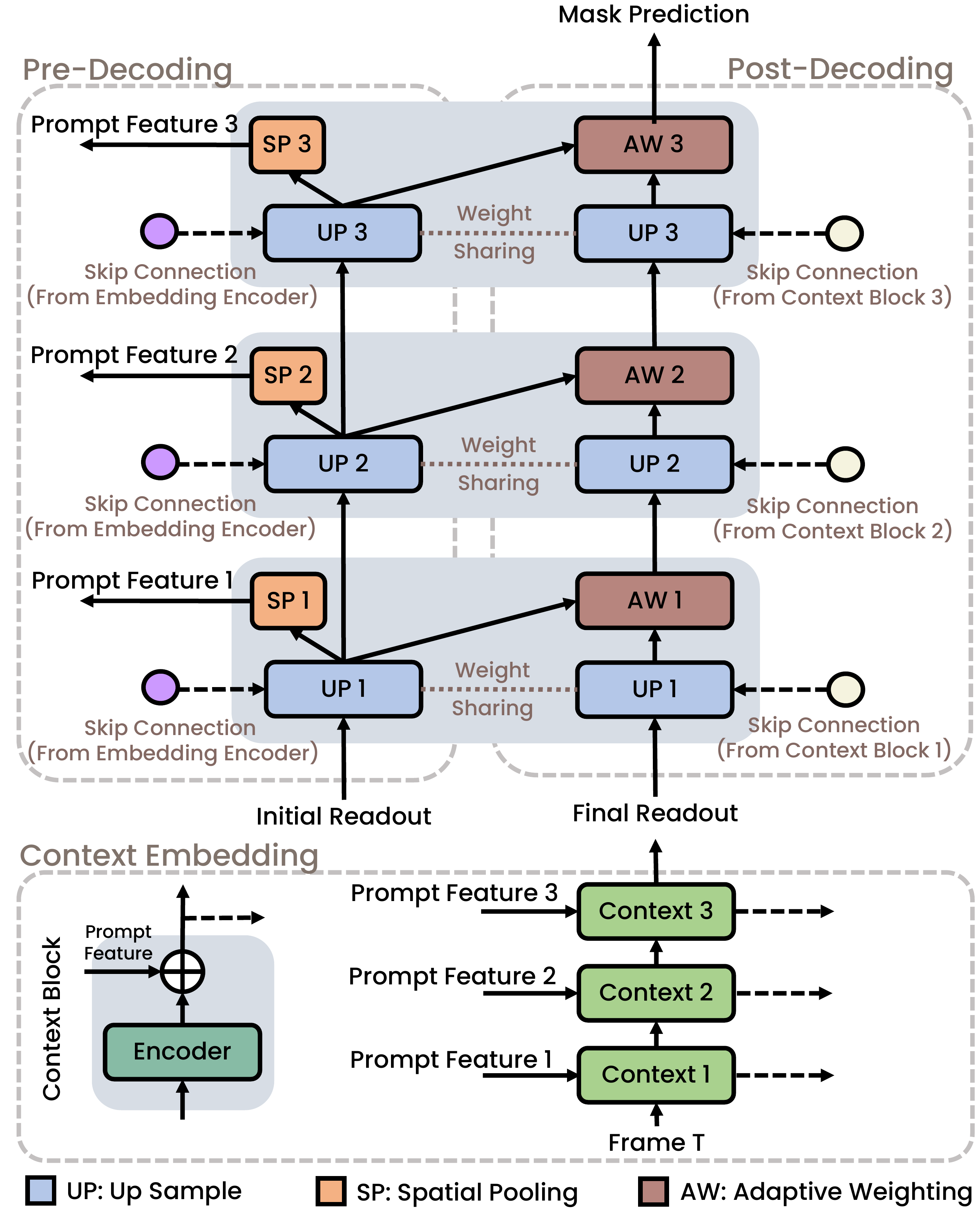}
    \caption{Illustration of compensatory decoding which compensates the low-level information for the initial readout.}
    \label{fig:cd}
\end{wrapfigure}

We observe that only improving the memory matching is insufficient in solving the object confusion and missing; thus, we propose a new compensatory decoding mechanism to further resolve the issues.
As shown in \ref{fig:cd}, the compensatory decoding process consists of three steps, 1) pre-decoding, 2) context embedding, and 3) post-decoding.
Pre-decoding aims to obtain a set of upsampled intermediate results as the guide spatial prompt in context embedding.
Context embedding gradually recovers the lost critical features in the initial memory readout $r_o$ by embedding the frame $T$ and spatial prompt feature, resulting
a final readout feature $r_o^{'}$.
In post-decoding, we generate the mask prediction base on the final readout feature $r_o^{'}$ and the skip-connections from context embedding.

\noindent\textbf{Context Embedding.}
Context block (CB) applies the same residual network layer as the encoder and three context blocks form a cascaded structure.
Prompt features $g_i$ and the residual encoder output $Res(f_{i-1})$ fuse as $f_i$ in each cascade:

\begin{footnotesize}
\begin{equation}
f_{i} = g_i + Res(f_{i-1}),
\end{equation}
\end{footnotesize}

where $Res$ is the residual layer.
$f_i$ is the input to the next CB as well as the skip connection for post-decoding, which contains richer semantic information than the query feature $e_q$ from embedding encoder.
Critical information of the target objects is sufficiently embedded into the final readout feature $r_o^{'}$, and the false matches in the initial memory readout are suppressed.

\noindent\textbf{Recursive Decoding Process.}
Inspired by \cite{DDRS}, we introduce the \textit{"looking and thinking twice"} idea into memory readout decoding, namely recursive decoding, which consists of pre-decoding and post-decoding.
Recursive decoding process shares weights in upsample blocks (\textbf{UP}, as shown in \ref{fig:cd}), which is essential for improving readout decoding due to seeing both the pre-decoded features and the context compensated features.
Let $D^b_{i}$ and $D^p_{i}$ denote pre-decoding and post-decoding outputs, $i$ is the cascading level.
The implementation of \textbf{UP} can be formally defined as follows:

\begin{footnotesize}
\begin{equation}
D^{p}_{i} = upsample(D^{p}_{i-1}, u) + f_i,
\end{equation}
\end{footnotesize}

where $u$ is the upsample rate, here $u$ is equal to 2.
Spatial pooling block (\textbf{SP}, as shown in \ref{fig:cd}) produce the spatial prompt feature in pre-decoding step, which is implemented via atrous spatial pyramid pooling block (ASPP).
In post-decoding step, we implement an adaptive weighting block (\textbf{AW}, as shown in \ref{fig:cd}) in each cascaded \textbf{UP} block to fuse context features and pre-decoding features.
We apply a convolution with a kernel size equal to 1 as the \textbf{AW} block.
The adaptive weighting process formulates as:

\begin{footnotesize}
\begin{equation}
w = sigmoid(\textbf{AW}(D^{p}_i)),
\end{equation}
\end{footnotesize}
\begin{footnotesize}
\begin{equation}
D_i = w * D^{p}_{i} + D^{b}_{i}*(1 - w),
\end{equation}
\end{footnotesize}
where $w$ is the weight of the post-decoding output.
Then $D_i$ is the input of the next cascaded \textbf{UP} block.

\begin{table*}[htb]
    \fontsize{4}{4.5}\selectfont 
    \begin{threeparttable}
        \resizebox{0.999\textwidth}{!}{
            \centering
            \begin{tabular}{lccc|cccc|ccccccc ccccc ccccc}
                \midrule
                ~                                                              & \multicolumn{3}{c}{\textbf{DAVIS 2017 test-dev}} & \multicolumn{4}{c}{\textbf{DAVIS 2017 val}} & \multicolumn{3}{c}{\textbf{DAVIS 2016 val}}                                                                                                                                                                                 \\
                \midrule
                \textbf{Method}                                                & $\mathcal{J}\&\mathcal{F}$                       & $\mathcal{J}$                               & $\mathcal{F}$                               & $\mathcal{J}\&\mathcal{F}$ & $\mathcal{J}$ & $\mathcal{F}$ & FPS                            & $\mathcal{J}\&\mathcal{F}$     & $\mathcal{J}$                  & $\mathcal{F}$ \\
                \midrule
                (CVPR$^{\prime}$2018) \textbf{RGMP} \cite{RGMP}               & 52.8                                             & 51.3                                        & 54.4                                        & 66.7                       & 64.8          & 68.6          & -                              & 68.8                           & 68.6                           & 68.9          \\
                (CVPR$^{\prime}$2019) \textbf{FEELVOS} \cite{FEELVOS}         & 57.8                                             & 55.1                                        & 60.4                                        & 71.6                       & 69.1          & 74.0          & 2.0                            & 81.7                           & 81.1                           & 82.2          \\
                (ICCV$^{\prime}$2019) \textbf{STM} \cite{STM}                 & 72.2                                             & 69.3                                        & 75.2                                        & 81.7                       & 79.2          & 84.3          & -                              & 89.4                           & 88.7                           & 90.1          \\
                (ECCV$^{\prime}$2020) \textbf{CFBI}  \cite{CFBI}              & 74.8                                             & 71.1                                        & 78.5                                        & 81.9                       & 79.1          & 84.6          & 5.9                            & 89.4                           & 88.3                           & 90.5          \\
                (CVPR$^{\prime}$2021) \textbf{RMNet} \cite{RMNet}             & 75.0                                             & 71.9                                        & 78.1                                        & 83.5                       & 81.0          & 86.0          & -                              & 88.8                           & 88.9                           & 88.7          \\
                (ECCV$^{\prime}$2020) \textbf{KMN} \cite{KMN}                 & 77.2                                             & 74.1                                        & 80.3                                        & 82.8                       & 80.0          & 85.6          & 4.2                            & 90.5                           & 89.5                           & 91.5          \\
                \textbf{CFBI+} \cite{CFBI2}                                   & 78.0                                             & 74.4                                        & 81.6                                        & 82.9                       & 80.1          & 85.7          & 5.6                            & 89.9                           & 88.7                           & 91.1          \\
                (ICCV$^{\prime}$2021) \textbf{HMMN} \cite{HMMN}               & 78.6                                             & 74.7                                        & 82.5                                        & 84.7                       & 81.9          & 87.5          & 10.0                           & 90.8                           & 89.6                           & 92.0          \\
                (NeurIPS$^{\prime}$2021) \textbf{AOT-L} \cite{AOT}            & 79.6                                             & 75.9                                        & 83.3                                        & 84.9                       & 82.3          & 87.5          & 18.0                           & 91.1                           & 90.1                           & 92.1          \\\
                (NeurIPS$^{\prime}$2021) \textbf{STCN} \cite{STCN}            & 79.9                                             & 76.3                                        & 83.5                                        & 85.3                       & 82.0          & 88.6          & 20.2                           & 91.7                           & 90.4                           & 93.0          \\
                \textbf{AOST-L} \cite{AOST}                                   & 79.9                                             & 76.2                                        & 83.6                                        & 85.6                       & 82.6          & 88.5          & 17.5                           & 92.1                           & 90.6                           & 93.6          \\
                (NeurIPS$^{\prime}$2022) \textbf{DeAOT-L} \cite{DeAOT}        & 80.7                                             & 76.9                                        & 84.5                                        & 85.2                       & 82.2          & 88.2          & 19.8                           & 92.3                           & 90.5                           & 94.0          \\
                (ECCV$^{\prime}$2022) \textbf{XMem} \textdagger \cite{XMem}   & 81.0                                             & 77.4                                        & 84.5                                        & 86.2                       & 82.9          & 89.5          & \textcolor{red}{\textbf{22.6}} & 91.5                           & 90.4                           & 92.7          \\
                (CVPR$^{\prime}$2023) \textbf{ISVOS} \textdagger \cite{ISVOS} & 82.8                                             & 79.3                                        & 86.2                                        & 87.1                       & 83.7          & 90.5          & -                              & \textcolor{red}{\textbf{92.6}} & \textcolor{red}{\textbf{91.5}} & 93.7          \\
                \midrule

                \textbf{JIMD}~(ours)                                & \textcolor{red}{\textbf{83.9}}                   & \textcolor{red}{\textbf{80.3}}              & \textcolor{red}{\textbf{87.4}}
                                                                               & \textcolor{red}{\textbf{88.1}}                   & \textcolor{red}{\textbf{85.2}}              & \textcolor{red}{\textbf{91.0}}              & 13.2
                                                                               & 92.4                                             & 90.6                                        & \textcolor{red}{\textbf{94.2}}                                                                                                                                                                                              \\
                \midrule
            \end{tabular}
        }
        \caption{\textbf{Comparison with state-of-the-art methods on DAVIS} (i.e., DAVIS 2017 test-dev set, DAVIS 2017 validation set and DAVIS 2016 validation set). \textbf{\textdagger}: Specific strategy without BL30K pre-training. AOT-L \cite{AOT} and its variants \cite{AOST,DeAOT} use the ResNet50 as backbone. The best score in each column is on \textcolor{red}{red} bold-faced.}
        \label{tab_sota}
    \end{threeparttable}
    \label{t1}
    \vspace{-20pt}
\end{table*}

\section{Experiments}
\subsection{Implementation Details}
\noindent\textbf{Training.}
We deploy the ResNet50 as the backbone for embedding encoder.
Following popular matching-based VOS methods \cite{STM,STCN,CFBI}, we employ the two-stage training strategy.
In the first training stage, the model is pre-trained on static datasets from AOT \cite{AOT}, where objects with masks are augmented (e.g., flip, shift, crop) and randomly synthesized onto the backgrounds.
In the second training stage, we perform training on real videos.
For the evaluation of DAVIS \cite{Davis17}, we use the training sets of DAVIS and YouTube, while for the evaluation of YouTubeVOS \cite{YT2018}, we only use the YouTube training set.
The embedding encoder is frozen in the second training stage to avoid overfitting to the seen object categories.
All modules apply the learning rates from initial 2e-4 then the learning rates gradually decay to 2e-5 in a polynomial manner.
We employ the cross entropy loss and soft jaccard loss \cite{JACCARDLOSS} to train the model. During training, we use an input size of $384\times 384$ and a batch size of 8, which is distributed on 4 RTX3090 GPUs.
Note that our method \textbf{do not} adopt the BL30K for training in the following reports.

\noindent\textbf{Inference.}
Our method uses a resolution of 480p by inference. Following common matching-based methods, we set the update frequency of memory to 5 (i.e., every five frames stack the short-term memory in long-term memory).
For a fair comparison, we \textbf{do not} employ the multi-scale inference trick on val/test datasets.

\subsection{Datasets and Metrics}
We evaluate our method on the five most popular VOS task benchmark datasets, consisting of a single-object dataset (DAVIS2016-val) and four multi-object datasets (DAVIS2017-val, DAVIS2017-test, YouTube2018-val, YouTube2019-val). A total of 971 real videos incorporated in the evaluation.
We evaluate our method by region similarity (i.e., $J$) and contour accuracy (i.e., $F$).
In YouTubeVOS, there are additional 26 unseen categories;
thus we separately report the $J$ score and the $F$ score for \textit{"seen classes"} in training set and \textit{"unseen classes"} that are not. $G$ is the global average score of all metrics.
We submit the val/test results on official online evaluation servers for a fair comparison.

\subsection{Compare with the State-of-the-Art Methods}

\noindent\textbf{Quantitative Comparison.}
As shown in Table \ref{tab_sota}, we compare the performance of our method with up-to-date methods on a series of DAVIS datasets.
Compared with the state-of-the-art memory matching-based methods (i.e., XMem \cite{XMem}), we achieve a 2.9\% and 1.9\% J\&F improvement on DAVIS2017 Test and DAVIS2017 Val, respectively.
Compared with the latest understanding-based method (i.e., ISVOS \cite{ISVOS}), we improve the J\&F by 1.1\% and 1\% on DAVIS2017 Test\&Val, respectively.
Our method achieves the top ranking of F performance on DAVIS 2016 single object performance, DAVIS2017 Test\&Val multiple objects performance.
On the Youtube dataset validation, for a fair comparison, we only adopt YouTubeVOS's train-set for training and compare it with XMem \cite{XMem} schemes using the same training data.
As shown in Table \ref{tab_ytb}, even without any fancy training and inference tricks, our method still achieves excellent performances in YouTube2018\&2019.

\noindent\textbf{Unseen Categories.}
Youtube dataset needs to evaluate the performance of unseen categories in training, as shown in Table \ref{tab_ytb}; our method outperforms other methods under the metrics of unseen categories.
Our proposed compensatory decoding stage generates guide information in pre-decoding that can more discriminatively help segment unseen category objects.

\begin{figure*}[t]
    \centering
    \includegraphics[width=\textwidth]{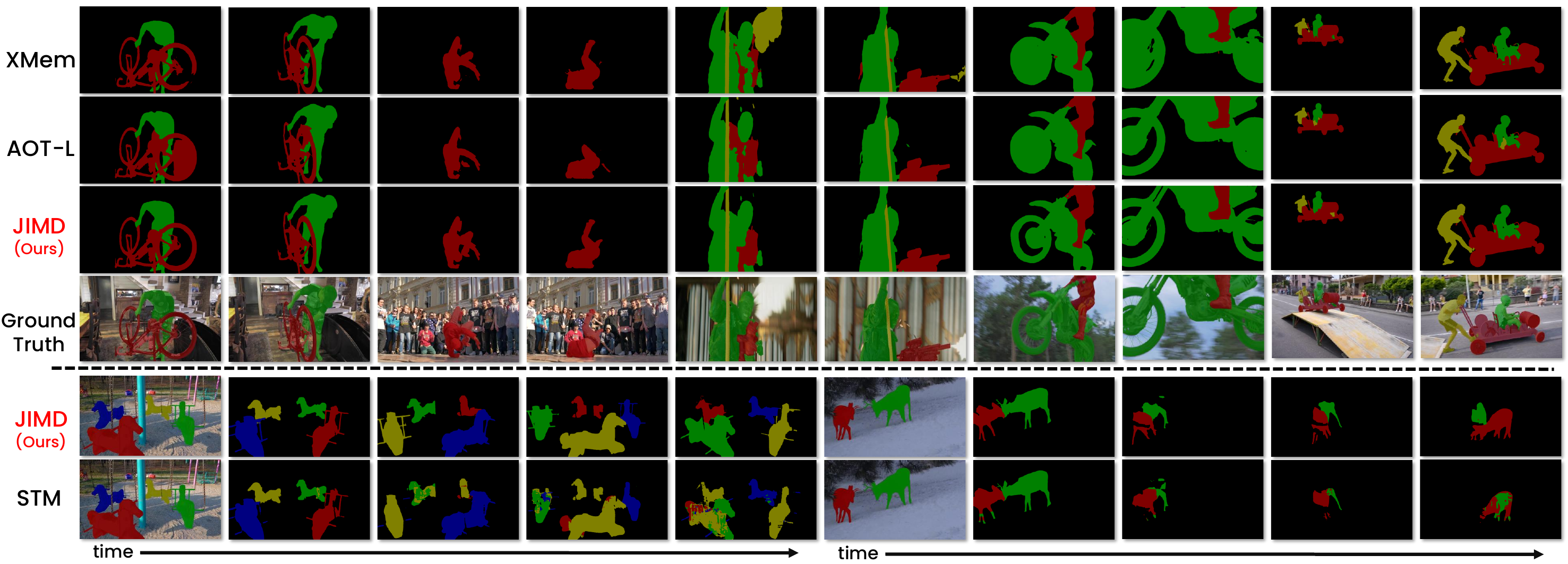}
    \caption{Representative challenge cases of qualitative comparison with XMem \cite{XMem}, AOT \cite{AOT}, and STM \cite{STM}.}
    \label{fig_qr}
\end{figure*}

\noindent\textbf{Qualitative Results.}
We visualize some video segmentation results in evaluation, including some common VOS challenge cases (i.e., tremendous motion, object reappearance, similar objects confusion), and compare them with two state-of-the-art matching-based methods AOT-L \cite{AOT} and XMem \cite{XMem}, as shown in Fig. \ref{fig_qr}.
Our method is superior to the other two methods in obtaining object details.
We can see that AOT-L \cite{AOT} and XMem \cite{XMem} make obvious errors when the human body appears in drastic motions.
XMem \cite{XMem} is more prone to errors when dealing with
objects of different scales (e.g., the rope and the human body in the Col 5 and 6 of Fig. \ref{fig_qr}).
Furthermore, we compare the classical method STM \cite{STM} based on memory matching for a long span.
Our method does not have object confusion after multiple similar objects are occluded, indicating that the proposed jointly improving method is helpful in overcoming the challenge of similar objects.

\subsection{Ablation Studies}
We improve the memory matching stage and the decoding stage separately, achieving cost-aware matching (CA) and cross-scale matching (CS) in the memory matching stage and compensatory decoding (CD) in the decoding stage.
In order to evaluate the effectiveness of the three improvements, we conduct separate experiments on JIMD for each modification
and explore the improvement performance of their combination.
All ablation studies are evaluated on DAVIS2017 Val split.
The corresponding modules used by the baseline in Table \ref{tab_ablation}
(i.e., long short-term matching modules, decoding process) are replaced by AOST-L \cite{AOST} (a evolution method of AOT-L \cite{AOT}), as shown in the first row of Table \ref{tab_ablation}.

\noindent
\textbf{Impact of Memory Matching Mechanism.}
As shown from Row 2 to Row 4 in Table \ref{tab_ablation},
both our cost-aware and cross-scale in the memory matching stage play positive roles when compensatory decoding is removed.
The improvement of cross-scale matching is more conducive to improving regional similarity (i.e., $J$), and cost-aware matching improves both metrics and edge accuracy (i.e., $F$).
Cross-scale matching has more powerful matching for objects of different scales in the video and thus is beneficial to improve region similarity.
Cost-aware constructs the cost volume for learning, explores the changes between two frames, and therefore is more effective
for preserving object details and edge features.
Cost-aware matching also improves the inference speed (FPS) in Col 7 of Table \ref{tab_ablation}, which abandons the calculation of neighbourhood cross-correlation in short-term memory and constructs the pixel relationship by dot product that is more efficient.
We can observe that the combined improvement strategy of cost-aware and cross-scale improves J by 2.6\% and F by 2\% compared to the baseline.

\begin{table}[t]
    \vspace{-1em}
    \hspace{-0pt}
    \begin{minipage}{0.55\textwidth}
        \begin{threeparttable}
            \fontsize{5.5}{0.5}\selectfont
            \scriptsize
            \begin{tabular}{lccccccccccc}
            \toprule
            \diagbox [] {\textbf{Methods}}{YTB2018} &$\bm{\mathcal{G}}$ & $\bm{\mathcal{J}_{s}}$ & $\bm{\mathcal{F}_{s}}$ & $\bm{\mathcal{J}_{u}}$ & $\bm{\mathcal{F}_{u}}$&\\
            \midrule
            CFBI                   & 81.4 & 81.1 & 85.8 & 75.3 & 83.4\\
            CFBI+                  & 82.8 & 81.8 & 86.6 & 77.1 & 85.6\\
            AOT-L$\star$                  & 84.1 & \textcolor{red}{\textbf{83.7}} & 88.5 & 78.1 & 86.1 \\
            STCN                   & 84.3 & 83.2 & 87.9 & 79.0 & 87.3 \\  
            XMem$\star$            & 84.4 & \textcolor{red}{\textbf{83.7}} & 88.5 & 78.2 & 87.2 \\
            \midrule
            \textbf{JIMD (ours)}$\star$ & \textcolor{red}{\textbf{84.8}} & \textcolor{red}{\textbf{83.7}} & \textcolor{red}{\textbf{88.7}} & \textcolor{red}{\textbf{79.1}} & \textcolor{red}{\textbf{87.6}}\\
            \midrule
            \midrule
            \diagbox [] {\textbf{Methods}}{YTB2019} &$\bm{\mathcal{G}}$ & $\bm{\mathcal{J}_{s}}$ & $\bm{\mathcal{F}_{s}}$ & $\bm{\mathcal{J}_{u}}$ & $\bm{\mathcal{F}_{u}}$&\\
            \midrule
            CFBI       & 81.0 & 80.6 & 85.1 & 75.2 & 83.0\\
            CFBI+      & 82.6 & 81.7 & 86.2 & 77.1 & 85.2\\
            AOT-L$\star$      & 84.1 & 83.5 & \textcolor{red}{\textbf{88.1}} & 78.4 & 86.3\\
            STCN       & 84.2 & 82.6 & 87.0 & 79.4 & 87.7 \\
            XMem$\star$  & 84.3 & \textcolor{red}{\textbf{83.6}} & 88.0 & 78.5 & 87.1 \\
            \midrule
            \textbf{JIMD (ours)}$\star$ & \textcolor{red}{\textbf{84.6}} & 82.9 & 87.8 & \textcolor{red}{\textbf{79.7}} & \textcolor{red}{\textbf{87.9}}\\
            \bottomrule
            \end{tabular}
            \caption{Results of YouTube2018\&2019 validation. $\star$: Stage2 only training by YouTube dataset (without extra data).}
            \label{tab_ytb}
            \end{threeparttable}
    \end{minipage}
    \begin{minipage}{0.4\textwidth}
            \begin{threeparttable}
            \vspace{-5em}
            \fontsize{9}{7}\selectfont
            \begin{tabular}{lccccccccccc}
            \toprule
            
            \textbf{CD} & \textbf{CS} & \textbf{CA} &$\mathcal{J\&F}$ & $\mathcal{J}$ & $\mathcal{F}$ & FPS\\
            \midrule
            \ding{55}       &\ding{55}       & \ding{55}          & 85.6 & 82.6 & 88.5 & 17.5\\
            \midrule
            \ding{55}       &\ding{51}       & \ding{55}          & 86.1 & 83.5 & 88.6 & 15.7\\
            \ding{55}       &\ding{55}       & \ding{51}          & 87.3 & 84.3 & 90.3 & \textcolor{red}{\textbf{17.8}}\\
            \ding{55}       &\ding{51}       & \ding{51}          & 87.7 & 84.8 & 90.5 & 14.3\\
            \midrule
            \ding{51}       &\ding{55}       & \ding{55}          & 87.6 & 84.9 & 90.2 & 17.1\\
            \ding{51}       &\ding{51}       & \ding{55}          & 87.7 & 85.0 & 90.3 & 13.6\\ 
            \ding{51}       &\ding{55}       & \ding{51}          & 87.9 & 85.0 & 90.8 & 15.0\\
            \ding{51}       &\ding{51}       & \ding{51}          & \textcolor{red}{\textbf{88.1}} & \textcolor{red}{\textbf{85.2}} & \textcolor{red}{\textbf{91.0}} & 13.2\\
            \bottomrule
            
            \end{tabular}
            \caption{Ablation performance on DAVIS 2017 Val of proposed improvements. CA: cost-aware matching. CS: cross-scale matching. CD: compensatory decoding.}
            \label{tab_ablation}
            \end{threeparttable}            
            
            
    \end{minipage}
\vspace{-2em}
\end{table}
\begin{table}[t]
    \vspace{-1em}
    \hspace{-0pt}
    \begin{minipage}{0.5\textwidth}
        \begin{threeparttable}
            \fontsize{9}{6.5}\selectfont
            \scriptsize
            \begin{tabular}{lccccccccccc}
            
            \toprule
            Method &$\mathcal{J\&F}$ & $\mathcal{J}$ & $\mathcal{F}$\\
            \midrule
            CD-Basline               & 87.6 & 84.9 & 90.2\\
            \hline
            \dtb{CD} {w/o Context}         & 86.4($\pmb{\downarrow}$\textbf{1.2})  & 83.5 ($\pmb{\downarrow}$\textbf{1.4}) & 89.3 ($\pmb{\downarrow}$\textbf{0.9})\\
            \hline
            CD w/o AW                & 87.4 & 84.6 & 90.1 \\
            \hline
            CD w/o SP                & 87.1 & 84.6 & 89.6 \\
            \bottomrule
            
            \end{tabular}
            \caption{Internal functional ablation for compensatory decoding.}
            \label{tab_decoding_modules}
            \end{threeparttable}
    \end{minipage}
    \begin{minipage}{0.5\textwidth}
        \begin{threeparttable}
            \scriptsize
            \begin{tabular}{lcccccccccccccc}
            
            \toprule
            \diagbox [] {Method}{$\mathcal{J\&F}$} & DAVIS 2017 & DAVIS 2016\\
            \midrule
            STM \cite{STM}                     & 81.7 & 89.4\\
            STM w/ CD              & 83.1 ($\pmb{\uparrow}$ \textbf{1.4}) & 90.0 ($\pmb{\uparrow}$ \textbf{1.2})\\
            \midrule
            TransVOS \cite{TransVOS}                 & 83.9 & 90.5\\
            TransVOS w/ CD         & 85.1 ($\pmb{\uparrow}$ \textbf{1.2}) & 90.7 ($\pmb{\uparrow}$ \textbf{0.2})\\
            \bottomrule
            
            \end{tabular}
            \caption{Compensatory decoding migrating to existing matching-based VOS methods.}
            \label{tab_decoding2others}
            \end{threeparttable}
    \end{minipage}
\vspace{-2em}
\end{table}

\noindent\textbf{Impact of Decoding Mechanism.}
Compensatory decoding provides an essential information supplement for matching readout results and suppresses false matching to a certain extent by implementing context embedding compensation in the decoding stage.
Row 5 of Table \ref{tab_ablation} demonstrates the improvement of compensatory decoding over the baseline.
To study the binding function effect in compensatory decoding, as shown in Table \ref{tab_decoding_modules}, we remove three implements or modules (i.e., context compensation (Context), adaptive weighting (AW), and spatial block (SP)) and evaluate the gains separately.
CD-Baseline in Table \ref{tab_decoding_modules} for the experimental setup in the 5th row of Table \ref{tab_ablation}.
On the context removal setting, we simultaneously remove the share-weighted recursive decoder and replace it with two decoders in the cascade that do not share weights.
As shown in Table \ref{tab_decoding_modules}, after removing context comparison, we can observe the most significant drop in performance (i.e., 1.4\% drop in $J$ and 0.9\% drop in $F$).
Therefore, context compensation is vital in improving the decoding stage.
Furthermore, we migrate our decoding improved mechanism to other existing matching-based methods, as shown in Table \ref{tab_decoding2others}, which indicates the feasibility and the plug-and-play potential of our compensatory decoding.

\noindent\textbf{Impact of Matching and Decoding Improved Jointly.}
The contribution of this paper is to explore the role of joint improvement of memory matching and decoding.
Rows 6 to 8 from Table \ref{tab_ablation} show the gain of the matching and decoding jointly improving mechanisms.
We can see that the combination of either CD \& CA or CD \& CS has a more substantial positive effect than the memory matching improved alone.
Therefore, we believe the joint improvement of memory matching and decoding proposed in this paper is crucial to the matching-based VOS approach.

\begin{figure}[h]
    \centering
    \vspace{-10pt}
    \setlength{\abovecaptionskip}{-0.cm}
    \includegraphics[width=\linewidth]{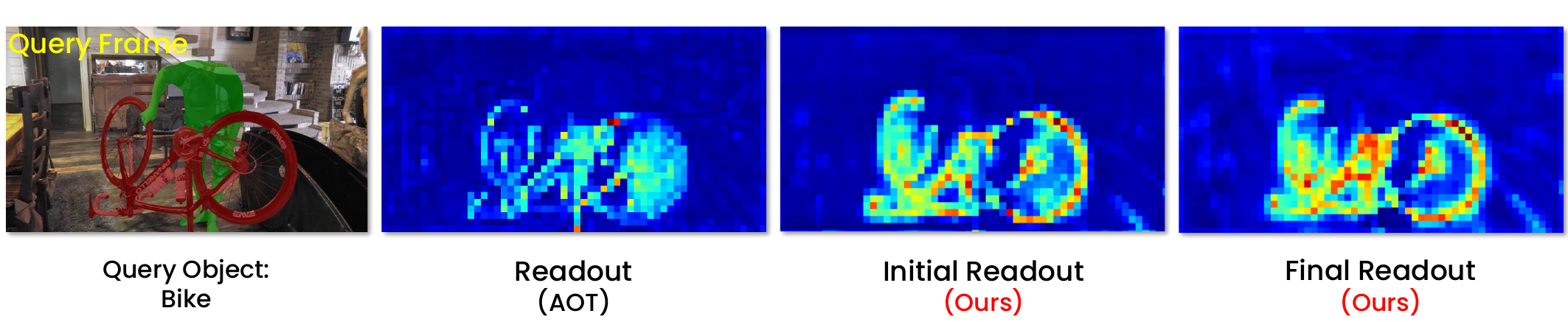}
    \caption{Visualization of the memory readout features of AOT, our method's initial readout and the final readout (i.e., feature after context embedding).}
    \label{fig_visz}
\end{figure}
\vspace{-10pt}

\noindent\textbf{Visualize the Readout Features.}
We visually compared the readout features with AOT \cite{AOT} as shown in Fig. \ref{fig_visz}.
After cost-aware and cross-scale matching, our initial readout results are significantly better than AOT \cite{AOT}.
We can see that our initial readout still has false matching area at the right wheel of the bike.
However, after context embedding, the final readout results suppress false matches and increase some high-response features.
This suggests that our approach of jointly improving the matching and decoding stages could facilitate producing more accurate and precise masks.

\vspace{-10pt}   
\section{Conclusion}
This paper proposes a network JIMD that jointly improves the memory matching and decoding stages to address the issue of false matching.
We design an improved mechanism for the memory matching stage consisting of cost-aware matching and cross-scale matching for short-term and long-term memory.
Cost-aware matching in short-term memory prompts the network to perceive changes between two frames more adequately.
Cross-scale matching in long-term memory prompts the network to explore the variations in different scaled objects.
For the readout decoding stage, we propose a novel compensatory decoding mechanism that can suppress false matches and supplement the crucial information loss of target objects.
We conduct extensive experiments on the effectiveness of joint improvement, and results on popular benchmarks demonstrate that JIMD outperforms existing matching-based methods.
Therefore, JIMD has considerable potential to be applied to multimedia applications in the future. 

\footnotesize
\noindent
\textbf{Acknowledgments.} This project was supported by the key R\&D project of Guangzhou (202206010091) and the fund of Southern Marine Science and Engineering Guangdong Laboratory (Zhanjiang)(ZJW-2023-04).

\bibliographystyle{splncs04.bst}
\bibliography{ref.bib}
\end{document}